\documentclass[times,twocolumn,final]{elsarticle}
\makeatletter\if@twocolumn\PassOptionsToPackage{switch}{lineno}\else\fi\makeatother

\usepackage{amsmath}
\usepackage{amsfonts}
\usepackage{mathtools}
\usepackage[caption=false,font=normalsize,labelfont=sf,textfont=sf]{subfig}
\usepackage{multirow}
\usepackage{bm}
\usepackage{algorithm}
\usepackage{algorithmic}
\usepackage{color, colortbl}
\definecolor{Gray}{gray}{0.9}
\definecolor{LightCyan}{rgb}{0.88,1,1}

\begin{document}

\begin{frontmatter}

\title{Multicriteria decision support employing adaptive prediction in a tensor-based feature representation}


\author[mymainaddress]{Betania Silva Carneiro Campello \fnref{myfootnote} \corref{mycorrespondingauthor}}
\author[secondadress]{Leonardo Tomazeli Duarte}
\author[mymainaddress]{Jo\~ao Marcos Travassos Romano}

\address[mymainaddress]{School of Electrical and Computer Engineering (FEEC), University of Campinas (UNICAMP), Campinas SP, Brazil}

\address[secondadress]{School of Applied Sciences (FCA), University of Campinas (UNICAMP), Limeira SP, Brazil}

\fntext[myfootnote]{E-mail addresses: betania@decom.fee.unicamp.br (B.S.C. Campello), leonardo.duarte@fca.unicamp.br (L.T. Duarte), romano@dmo.fee.unicamp.br (J.M.T. Romano)}

\fntext[myfootnote]{This work was supported by the National Council for Scientific and Technological Development (CNPq, Brazil), grant numbers, 168968/2018-5, 312228/2020-1.This work was also supported by the Sao Paulo Research  Foundation (FAPESP), grant 2023/04159-6, grant 2020/01089-9 and grant 2020/09838-0 (BI0S - Brazilian Institute of Data Science).}

%
\cortext[mycorrespondingauthor]{betania@decom.fee.unicamp.br}

\begin{abstract}
Multicriteria decision analysis (MCDA) is a widely used tool to support decisions in which a set of alternatives should be ranked or classified based on multiple criteria. Recent studies in MCDA have shown the relevance of considering not only current evaluations of each criterion but also past data. Past-data-based approaches carry new challenges, especially in time-varying environments. This study deals with this challenge via essential tools of signal processing, such as tensorial representations and adaptive prediction. More specifically, we structure the criteria' past data as a tensor and, by applying adaptive prediction, we compose signals with these prediction values of the criteria. Besides, we transform the prediction in the time domain into a most favorable decision making domain, called the feature domain. We present a novel extension of the MCDA method PROMETHEE II, aimed at addressing the tensor in the feature domain to obtain a ranking of alternatives.  Numerical experiments were performed using real-world time series, and our approach is compared with other existing strategies. The results highlight the relevance and efficiency of our proposal, especially for nonstationary time series.
\end{abstract}


\end{frontmatter}


\section{Introduction} 
\label{sec:intro}

Signal processing (SP) methods are widely used in many areas, including application fields such as acoustics, and biomedical data analysis. In contrast to these well-established applications, we address the use of SP methods in decision science, a field for which the potentials of SP methods have not yet been fully exploited. Indeed, there is an obvious room for consider SP-based approaches in decision-related tasks, which has been atested by several works which consider different aspects of the decision process. For instance, the least-mean-square algorithm was employed in decision making related to the COVID-19 pandemic in Italy~\cite{marano2022decision}.

In this study, we aim to exploit the potentialities of SP methods in a particular subfield of decision sciences, known as multiple-criteria decision analysis (MCDA)~\cite{hwang1981methods}. An important task in MCDA is to rank a set of alternatives based on multiple criteria. The input data of MCDA methods is usually a \textit{decision matrix}, in which the rows are associated with a set of alternatives, and the columns with a set of criteria. Each element of this matrix corresponds to the evaluation of an alternative in a given criterion. An illustrative example of an MCDA problem is to ranking some countries for deciding where to open a branch office considering as decision criteria the per capita income (PCI) and the purchasing index of the product (PI). The alternatives are the countries, and criteria 1 and 2 correspond to the PCI and PI values, respectively.

Most MCDA methods assume a single value for each criterion, which may be the average of the criterion performance in a given period, the latest available data of the criterion value, or some static data~\cite{martins2021multidimensional, campello2023exploiting}. However, several decisions are made considering the evolution of the criteria over time (their time series or signals) or more than one criteria' attribute simultaneously (e.g., their average and tendency). This statement is supported by several studies, such as~\cite{zhang2017signal}, which highlight the importance of time series analysis in economics; \cite{agrahari2021prognosticating} in socio-political, and \cite{dash2021intelligent} in healthcare decision-making.

Despite the relevance of temporal analysis in decision-making, few studies in MCDA have focused on time series-based approaches, as shown in~\cite{campello2023exploiting, martins2021multidimensional, kandakoglu2019multicriteria}. Examples of time-based approaches in MCDA include the studies by~\cite{frini2019mupom, mouhib2021tsmaa}, which proposed a methodology for dealing with time series in sustainable contexts, and~\cite{witt2021multi}, which applied multi-period analysis in a case study from the German energy sector. Other recent studies that deal with time series analysis in MCDA are those by~\cite{wkatrobski2017temporal, banamar2019interpolation, martins2021multidimensional}.

In~\cite{campello2020adaptive, campello2023exploiting, campello2022dealing} we extended the classical decision matrix to a tensorial formulation~\cite{kilmer2011factorization, chen2018tensor, chen2020multi}. The extension allows a representation in which one takes into account a temporal evaluation of the decision data. The resulting decision tensor comprises three dimensions that represent the alternatives, the criteria, and the time index. Having defined the decision tensor, we applied in~\cite{campello2020adaptive} the Normalized Least-mean-square (NLMS) and Recursive Least Square (RLS) algorithms to predict future criteria values, structuring these future values into a decision matrix. Then, we applied PROMETHEE II~\cite{brans2005promethee}, a well-established MCDA method, to obtain a ranking of alternatives. Alternatively, in \cite{campello2023exploiting}, we proposed an approach where the decision tensor of past time series is mapped to a tensor of time series features, such as trend, variance, and average. Thus, we proposed an extension of an MCDA method, known as TOPSIS~\cite{hwang1981methods, fu2023re}, to obtain a ranking of alternatives from these features. The results in both strategies showed that time-series-based approach brings valuable information to the decision-making process.

In this study, we propose a novel methodology that uses the decision tensor of past time series to predict future time series of the criteria. The predicted data is structured in a tensor, from which features are extracted. These data are formulated as a third-order tensor, where the dimensions represent the alternatives, the criteria, and the features of the predicted time series. To obtain the final ranking of the alternatives, we introduce a new extension of the PROMETHEE II method.

This work differs from previous papers that have explored time series in MCDA, as they obtain the ranking in two steps. First, a classical MCDA method is applied at each period of time. Then, again, the classical MCDA method is used to obtain a final ranking, as revised and highlighted in the study~\cite{campello2023exploiting}. In contrast, our approach aims to predict future criteria values and extract features. Furthermore, in the study~\cite{campello2023exploiting}, we utilize features from past data, and the resulting ranking may differ from that obtained using predicted data. This new approach can be particularly useful for decision-making when the criteria features are relevant and their time series are non-stationary.

Besides, to our knowledge, the only study that has proposed  a method for dealing with tensors is~\cite{campello2023exploiting}, which utilized an extension of TOPSIS. Therefore, it is relevant to develop  additional techniques for handling tensors, as different MCDA methods can yield distinct results for the same problem, and the method chosen should be appropriately selected based on the context of the problem~\cite{guitouni1998tentative}. In this study, we introduce an extension of the PROMETHEE II method, expanding its applicability to tensors and offering a novel methodological contribution.

\textcolor{red}{It is worth to highlight that, in the field of decision analysis, unlike, for instance, supervised machine learning, determining the superiority of methods is difficult as there is no access to ground truth. Considering this perspective, it is crucial to emphasize that the choice of a particular decision method should take into account the characteristics of the decision problem \cite{wkatrobski2016guideline, campello2023exploiting}. Our proposal is specifically tailored to exploit the decision space that arises when temporal characteristics are available.}

The paper is organized as follows. Section~\ref{sec:problemstatement} describes the MCDA problem and the motivation of the study . Section~\ref{sec:met} introduces the proposed methodology. Section~\ref{sec:numerical_experiments} provides a set of numerical experiments. Section~\ref{sec:conclusion} concludes this study.

\section{Problem statement and motivation}
\label{sec:problemstatement}

MCDA methods are widely used to rank alternatives according to several criteria. In the classical MCDA approaches, the data are modeling as a decision matrix, $\textbf{H} \in \mathbb{R}^{n \times m}$. The rows of $\textbf{H}$ are associated with a set of $n$ alternatives, $A = \{a_1, a_2, \cdots, a_n\}$, and the columns, with $m$ criteria, $C = \{c_1, c_2, \cdots, c_m\}$. The elements of this matrix are  $h_{ij}$, representing the evaluation of alternative $i$ in criterion $j$. 

Most MCDA approaches relies on the concept of \textit{matrix aggregation}. In brief, the aggregation consists of mapping the decision matrix $\textbf{H} \in \mathbb{R}^{n \times m}$ into a scoring vector $\textbf{f} \in \mathbb{R}^{n}$. Thus, each row $i$ (alternative $i$) of the matrix \textbf{H} is mapped into a score $f_i$, used to rank the alternatives. There are many aggregation methods available \cite{zopounidis2002multicriteria}, each with different characteristics that can be useful for different types of decision-making \cite{guitouni1998tentative}.

A well-established MCDA aggregation technique is the Preference Ranking Organization Method for Enrichment Evaluation (PROMETHEE) family method~\cite{brans2005promethee}. The preference structure used in this approach is based on a set of  \textit{pairwise comparisons}. It compares the evaluations of two alternatives on a given criterion, for all pairs of alternatives in all criteria. A weight vector $\boldsymbol{\Gamma}$ is considered to model the criteria' relative importance. The PROMETHEE II~\cite{brans2005promethee} technique uses as input the decision matrix \textbf{H}, and the output is the vector $ \textbf{f}  \in \mathbb{R}^n $ containing the aggregated scores. 

Many MCDA procedures statically use a criteria evaluation without considering their temporal evaluations. This study proposes to extend the classic static MCDA approach to a dynamic evaluation of the performance of alternatives. This dynamic approach should be made by evaluating the criteria' characteristics of the future time series.

\begin{figure}
	\begin{minipage}[b]{1.0\linewidth}
		\centering
		\centerline{\includegraphics[width=8cm]{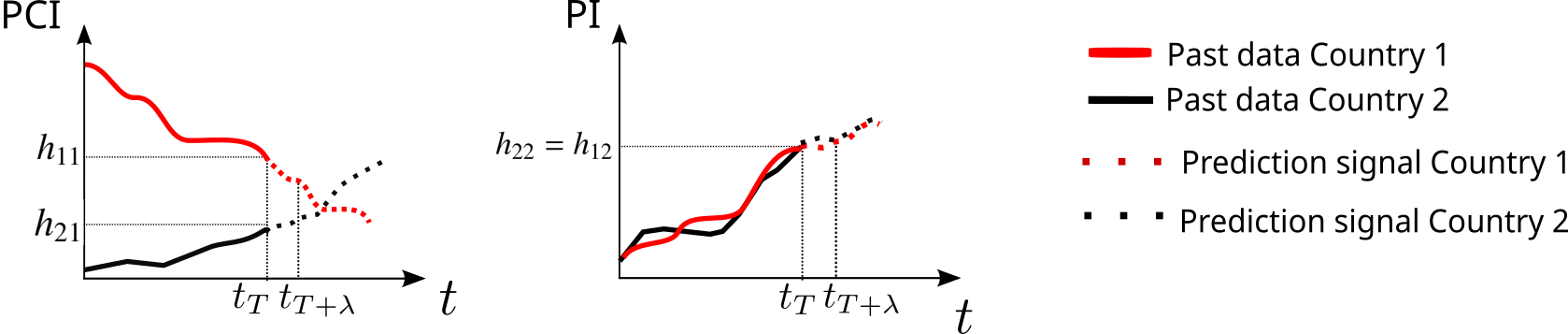}}
	\end{minipage}\caption{An example of opening a branch office in which the criteria prediction signals are considered.}
	\label{fig:grafico}
	\hfill
\end{figure}

The motivation of our approach can be seen in the branch office example given in Section~\ref{sec:intro}. In this example, the decision maker needs  to rank two countries according to  PCI and PI criteria. Figure~\ref{fig:grafico} shows the past signals and the predicted signals of both PCI and PI. Based on the current data  at $t_T$, we can conclude that Country 1 is a more favorable option than Country 2, as it has a higher PCI compared to Country 2 (i.e., $h_{11} > h_{21}$) while having the same PI as Country 2 (i.e., $h_{22} = h_{12}$). Country 1 may also be the more favorable option using approaches that aggregate time-series of criteria, as all PCI time-series values of Country 1 are higher than those of Country 2. Thus, if one only considers the data at instant time $ t_T $ or the time series of the criteria, then one misses important information related to the downward trend of the Country 1 PCI as well as the upward trend of the Country 2 PCI. 

Let us now consider the approach proposed in~\cite{campello2020adaptive}, which consists in applying some point prediction method in each signal of the PCI and PI criteria, in order to obtain the criteria predicted values for the period $t_{T+\lambda}$ (being $\lambda$ the prediction step). At period $t_{T+\lambda}$, $h_{11} > h_{21}$ and $h_{22} = h_{12}$, indicating that Country 1 remains the more favorable alternative, despite a trend indicating that Country 2 may outperform Country 1. 

Finally, this approach can be compared with the one proposed by~\cite{campello2023exploiting}, where features were extracted from past data. It is noteworthy that the average PCI of Country 1 based on past data is greater than that of Country 2. However, the average prediction signal for both countries is similar. Therefore, the approach proposed by~\cite{campello2023exploiting} yields a different ranking compared to our approach, where Country 1 and Country 2 may have equal performance. 

Thus, according to the methodology employed in previous studies, Country 1 may dominate Country 2. However, our approach reveals that Country 2 can emerge as the preferred choice. This example serves to illustrate that disregarding the features of future signals can potentially lead to overlooking a more suitable solution, particularly when the decision-making horizon extends to the medium- or long-term and signals exhibit non-stationary behavior.

\section{METHODOLOGY}
\label{sec:met}

\begin{figure*}
	\begin{minipage}[b]{1.0\linewidth}
		\centering
		\centerline{\includegraphics[width=18cm]{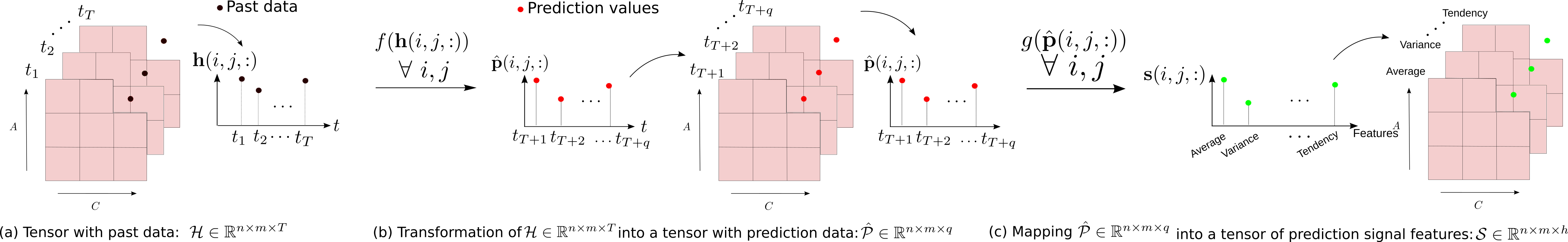}}
	\end{minipage}\caption{Tensorial proposal in which the features of the prediction signal are obtained from the past data.}
	\label{fig:esquema_proposta}
	\hfill
\end{figure*}

Figure~\ref{fig:esquema_proposta} provides an outline of our proposal. The input data corresponds to a tensor as represented in Figure~\ref{fig:esquema_proposta}(a). Given this tensor, the first step consists in predicting the values of each criterion over the next $q$-steps. These $q$ predicted values compose the prediction signal, $\hat{\textbf{p}}(i,j,:)$, as shown in Figure~\ref{fig:esquema_proposta}(a) -- (b), and described in Section~\ref{sec:predict_signal}. In a second step, described in Section~\ref{sec:mcda}, for each alternative $i$ and criterion $j$, we compute the time series features and structure them into a tensor of features, as illustrated in Figure~\ref{fig:esquema_proposta}(c). Finally, we propose an extension of the PROMETHEE II method to obtain the alternatives' ranking from the tensor of features. We adopted the PROMETHEE II technique since it presented satisfactory results in~\cite{campello2020adaptive}
using predicted values.

\subsection{Prediction step}
\label{sec:predict_signal}

In classic adaptive prediction, the problem of $\lambda$-steps prediction lies in establishing a mapping that, when applied to a set of samples of a time series, provides a predicted value $\lambda$-steps forward~\cite{romano2016unsupervised}. Given a time series $\textbf{h}(i, j, :)$ and its elements $h_{ijt}$, the input of the filter is a set of samples of $\textbf{h}(i, j, :)$, denoted by  $\textbf{h}_{ij}(t-\lambda) = [h_{ij(t-\lambda)}, \ldots, h_{ij(t-\lambda - M)}]^T$, where $ M $ is the filter size, and $t = 1,\ldots, T$. The filter output is
{\small
\begin{align}\label{eq:predictionvalue}
	\hat{p}_{ijt} = f\{\textbf{h}_{ij}(t-\lambda)\}.
\end{align} }

\noindent The mapping $f\{.\}$ is a combination of the past samples $\textbf{h}_{ij}(t-\lambda)$  with a set of $M$ parameters $\textbf{w} = [ w_1, \ldots, w_M]$. In a linear combination, Eq.~(\ref{eq:predictionvalue}) can be rewritten as $\hat{p}_{ijt} = \textbf{w}^T\textbf{h}_{ij}(t-\lambda)$, and the prediction at the instant $T +\lambda$ is:
{\small
\begin{align}\label{eq:predc} 
	\hat{p}_{ij(T + \lambda)} =   \textbf{w}_{T+1}^T \textbf{h}_{ij}(T)
\end{align} }

In this study, the prediction step is carried out by the RLS algorithm for two main reasons: the fast convergence rate with which RLS tends to the optimal solution; and the ability to adapt to dynamic changes in the data. Additionally, RLS is relatively easy to implement and requires a computational complexity compatible with our problem. The input data for the prediction step is the decision tensor $\mathcal{H}  \in \mathbb{R}^{n \times m \times T}$, where $ n $, $ m $ and $ T $ denote the index associated with the alternatives, criteria and time samples, respectively, and by using prediction steps $\lambda = 1,\ldots, q$, the output is $\hat{\mathcal{P}}  \in \mathbb{R}^{n \times m \times q}$. 

\subsection{PROMETHEE II extension for a tensor}
\label{sec:mcda}

PROMETHEE is considered an outranking-based approach, in which the preference structure is computed by means of pairwise comparisons between alternatives. Due to the nature of its technique, it is not necessary to normalize the input data, which is an advantage as it can avoid round-off errors that may occur during normalization~\cite{feng2020enhancing}. Additionally, PROMETHEE has proven to be useful in many real-world applications~\cite{behzadian2010promethee}. Because of the strength of PROMETHEE in both conception and application, we propose in this study a new variant of PROMETHEE II, in which the input is a tensor $ \hat{\mathcal{P}}  \in \mathbb{R}^{n \times m \times q}$ (obtained in the prediction step) and the output is the vector $\hat{\textbf{f}}$, which represents an aggregated scores. 

In the first step, the features of each fiber $\hat{\textbf{p}}(i,j,:)$, is computed, mapping the time-space into the feature space, as:
\vspace{-0.2cm}
{\small
\begin{eqnarray}\label{eq:transf_atributo}
	\hat{\mathcal{P}}  \in \mathbb{R}^{n \times m \times q} \Rightarrow \mathcal{S}  \in \mathbb{R}^{n \times m \times w},
\end{eqnarray} } 
\vspace{-0.55cm}

\noindent where $w$ is the number of features. In our case, we consider, without lost of generality, three features: average, slope coefficient (S.C.), and coefficient of variation (C.V.). 

In the next step, for each slice $S(:,:,\ell)$, $\ell = 1,\ldots, w$ a pairwise comparisons is made by compute the difference between the performance of each couple of alternative $a_i, \ a_k \in A$, for each criterion $j$:
\vspace{-0.2cm}
{\small
\begin{align}\label{eq:parwise} 
	d_{j\ell}(a_i, a_k) = h_{ij\ell} - h_{kj\ell} \ \ \forall  j, \ell.
\end{align}}

From~(\ref{eq:parwise}), we obtain a tensor $\mathcal{D}  \in \mathbb{R}^{n \times n \times m \times w}$. Each slice $\textbf{D}(:,:,j,\ell) \in \mathbb{R}^{n \times n}$ of $\mathcal{D}$ is a pairwise comparison matrix.   

Afterward, a function is applied, called preference function $P$, in $\textbf{D}(:,:,j,\ell)$ as: 
{\small
\begin{align}\label{eq:p}
	P_{j\ell}(d_{j\ell}(a_i, a_k)) = P_{j\ell}(\textbf{D}(:,:,j,\ell))  \ \ \forall j ,\ell.
\end{align} }
where $0 \leq P_{j\ell}(\textbf{D}(:,:,j,\ell)) \leq 1$. The primary preference function~\cite{brans2005promethee} is used, given by $P_{j\ell}(d_{j\ell}(a_i, a_k)) = \dfrac{\mbox{sgn}(d_{j\ell}(a_i, a_k))+1}{2}$, where sgn($ d_{j\ell}(a_i, a_k)) $ represents the sign function.

By considering a set of $m \times w$ weights $\boldsymbol{\Gamma}_\ell	 = \{\gamma_{1\ell}, \gamma_{2\ell}, \cdots,$ $\gamma_{m\ell}\}$; where $\gamma_{j\ell} \geq 0 \ \ \forall j, \ell$, and $\sum_{\ell=1}^{w}\sum_{j=1}^{m}$ $\gamma_{j\ell} = 1$, the next step mapping  $\mathbb{R}^{n \times n \times m \times w} \rightarrow \mathbb{R}^{n \times n}$, which provides a global preference index of $a_i$ over $a_k$ $\forall$ $i$, $k$:
{\footnotesize
\begin{align}\label{eq:pi}
	\small
	\pi(a_i, a_k) = \sum_{\ell=1}^{w} \sum_{j=1}^{m}\gamma_{j\ell} P_{j\ell}(\textbf{D}(:,:,j,\ell)).
\end{align} }
Since $P_{j\ell}(\textbf{D}(:,:,j,\ell)) \geq 0$ and $\gamma_{j\ell} \geq 0$, then $\pi(a_i, a_k) \geq 0$.

The final step is an aggregation process that maps $  \pi(a_i, a_k) $ to a vector $ \hat{\textbf{f}} $. This process is given computing the difference  between the mean preference of $a_i$ over the other alternatives, and the mean preference of all alternatives over $a_i$:
{\footnotesize
\begin{align}\label{eq:ranking}
	\hat{\textbf{f}} = \dfrac{1}{n-1}\sum_{a \in A}\pi(a, a_i) - \dfrac{1}{n-1}\sum_{a \in A}\pi(a_i, a), \ \ \forall i.
\end{align} }
Each element of vector $\hat{\textbf{f}}$, which lies in the interval $[-1,1]$, allows one to ranking the alternatives.

\section{Numerical experiments}
\label{sec:numerical_experiments}

In this section the proposed approach is tested by considering real time series taken from the International Monetary Fund (IMF): https://www.imf.org/en/Data. We provide these time series, their respective graphs, and the Python code on the link https://github.com/BSCCampello/tensorpredictionsignals. 

Let us suppose the decision maker wants to rank five countries ($n=5$): $a_1$ -- Belgium; $a_2$ -- Canada; $a_3$ -- France; $a_4$ -- Japan; $a_5$ -- Netherlands,  according to three financial-economic criteria ($ m=3 $): gross national savings ($c_1$), inflation ($c_2$), and unemployment rate ($c_3$). The criterion $c_1$ is considered of maximum, the criteria  $c_2$ and $c_3$ of minimum. The feature given by the coefficient of variation is expected to be minimized, as it is related to risk. The available time series consists of 39 samples ($T = 39$): $ t =1980, \ldots, 2018$. 

To minimize prediction error, we performed preliminary adjustments before determining the parameters of the RLS algorithm obtaining: the forgetting factor was 0.99 for signals of criterion 2 and 0.90 for signals of criteria 1 and 3. Additionally, we used two parameters ($\textbf{w}$). The weights of the PROMETHEE II criteria were determined to be equal for all criteria: $\mathbf{\gamma}_{i\ell} = \frac{1}{9}, \ \forall i, \ell$. 

\begin{table}
	\caption{Comparison of rankings obtained from three different strategies.}\label{tab:ranks}
	\begin{center}
		\scalebox{0.7}{%
			\begin{tabular}{clccccc}
				\hline
				Year&Ranking&1st &2nd &3rd &4th &5th \\\hline\hline
				\multirow{2}{*}{Features}&Benchmark $\textbf{f}^*$		&$ a_5$& $ a_4 $&$ a_3 $&$ a_1 $&$ a_2$\\
				2013--2018&Prediction $\hat{\textbf{f}}$		&$ a_5$& $ a_4 $&$ a_3 $&$ a_1 $&$ a_2$\\\hline
				2012&Current $\textbf{g}^c$	&$ a_4$& $ a_5 $&$ a_1 $&$ a_2 $&$ a_3$\\\hline\hline
				\multirow{2}{*}{2013}&Reference $\textbf{g}^*$	&$a_4$& $ a_5 $&$ a_1 $&$ a_2 $&$ a_3$\\
				&Prediction $\hat{\textbf{g}}$	&$a_4$& $ a_5 $&$ a_1 $&$ a_2 $&$ a_3$\\\hline
				\multirow{2}{*}{2014}&Reference $\textbf{g}^*$	&$a_4$& $ a_5 $&$ a_1 $&$ a_2 $&$ a_3$\\
				&Prediction $\hat{\textbf{g}}$	&$a_4$& $ a_5 $&$ a_1 $&$ a_2 $&$ a_3$\\\hline
				\multirow{2}{*}{2015}&Reference $\textbf{g}^*$	&$a_5$& $ a_4 $&$ a_1 $&$ a_3 $&$ a_2$\\
				&Prediction $\hat{\textbf{g}}$		&$a_5$& $ a_4 $&$ a_1 $&$ a_2 $&$ a_3$\\\hline
				\multirow{2}{*}{2016}&Reference $\textbf{g}^*$	&$a_4$& $ a_5 $&$ a_1 $&$ a_3 $&$ a_2$\\
				&Prediction $\hat{\textbf{g}}$		&$ a_5$& $ a_4 $&$ a_1 $&$ a_2 $&$ a_3$\\\hline
				\multirow{2}{*}{2017}&Reference $\textbf{g}^*$	&$a_4$& $ a_5 $&$ a_1 $&$ a_3 $&$ a_2$\\
				&Prediction $\hat{\textbf{g}}$		&$ a_5$& $ a_4 $&$ a_1 $&$ a_3 $&$ a_2$\\\hline
				\multirow{2}{*}{2018}&Reference $\textbf{g}^*$	&$a_4$& $ a_5 $&$ a_1 $&$ a_3 $&$ a_2$\\
				&Prediction $\hat{\textbf{g}}$		&$ a_5$& $ a_4 $&$ a_1 $&$ a_3 $&$ a_2$\\\hline\hline
				Features&2007--2012 $\textbf{g}$		&$ a_5$& $ a_4 $&$ a_1 $&$ a_3 $&$ a_2$\\\hline
		\end{tabular}}
	\end{center}
\end{table}
\begin{table*}
	\caption{Feature tensor of the prediction signals, the matrix of current data, and feature tensor of the signals from 2007 to 2012.}\label{tab:valores_feature_predicao_e_current}
	\begin{center}
		\scalebox{0.6}{%
			\begin{tabular}{|c||ccc|ccc|ccc||ccc||ccc|ccc|ccc|}
				\hline
				\multirow{4}{*}{$a_i$}&\multicolumn{9}{c||}{Feature tensor of the prediction signals $\mathcal{S}$}&\multicolumn{3}{c||}{\textbf{H}}&\multicolumn{9}{c|}{Feature tensor of the past data signals $\mathcal{T}$}\\\hline
				&\multicolumn{3}{c|}{Average}&\multicolumn{3}{c|}{S.C.}&\multicolumn{3}{c||}{C.V.
				}&\multicolumn{3}{c||}{Current data}&\multicolumn{3}{c|}{Average}&\multicolumn{3}{c|}{S.C.}&\multicolumn{3}{c|}{C.V.}\\			
				&$c_1$ &$c_2$ &$c_3$ &$c_1$ &$c_2$ &$c_3$ &$c_1$ &$c_2$&$c_3$&$c_1$&$c_2$ &$c_3$&$c_1$ &$c_2$ &$c_3$ &$c_1$ &$c_2$ &$c_3$ &$c_1$ &$c_2$&$c_3$ \\
				&Max &Min &Min &Max &Min &Min &Min &Min &Min&Max &Min &Min
				&Max &Min &Min &Max &Min &Min &Min &Min &Min\\\hline\hline
				\rowcolor{Gray}$a_1$& 22.7 & 105.6 &  7.2& -0.069 &   2.255 & -0.114 &0.007 &   0.037 &  0.027& 23.15 & 97.68 &  7.55 &23.631 & 91.846 & 7.563&-0.443 &  2.154 & 0.035&0.072 &  0.041 & 0.060  \\\hline
				\rowcolor{LightCyan}$a_2$&20.9 & 131.5 &  7.0 &-0.003 &   2.905 & -0.078 &
				0.012 &   0.038 &  0.021&21.28& 121.68 &  7.32&21.733 & 116.333 & 7.229&-0.699 &   2.010 & 0.289&0.095 &   0.030 & 0.120  \\\hline
				\rowcolor{Gray}$a_3$&21.4 & 105.6 &  9.8&-0.054 &   2.110 & -0.050&0.006 &   0.034 &  0.011&21.66& 98.33 &  9.79 &22.298 & 93.813 & 8.798&-0.444 &  1.635 & 0.414&0.054 &  0.030 & 0.092 \\\hline
				\rowcolor{LightCyan}	$a_4$&22.4 & 97.8 & 4.4&-0.277 &  0.453 & 0.047&0.027 &  0.008 & 0.022&23.62 & 96.22 &  4.33 &25.610 & 97.006 & 4.476&-1.036 & -0.360 & 0.121 &0.079 &  0.008 & 0.107 \\\hline
				\rowcolor{Gray}	$a_5$&30.1 & 103.9 &  5.4&0.149 &   2.011 & -0.2&0.009 &   0.033 &  0.113&29.39& 96.98 &  5.83&28.011 & 92.203 & 4.666&0.274 &  1.598 & 0.370&0.045 &  0.030 & 0.150\\\hline
		\end{tabular}}
	\end{center}
\end{table*} 
We consider a situation in which the decision should be made in 2012 (i.e., suppose the time series are available from 1980 until 2012); thus, we can use data for 2013 to 2018 as benchmark for assessing the proposed strategy. Therefore, the ranking provided by the proposal (represented by $\hat{\textbf{f}}$) is obtained by considering the tensor $ \mathcal{H} \in \mathbb{R}^{n \times m \times T}$, where $n = 5$, $m = 3$, $T = 33$ ($t = 1980, $\ldots$, 2012$), as input for the prediction step. The output is a tensor of predicted values $\hat{\mathcal{P}} \in \mathbb{R}^{5 \times 3 \times q}$, where $q = 6$, i.e., the prediction is for $ t = 2013, \ldots, 2018$. The tensor  $\hat{\mathcal{P}} \in \mathbb{R}^{5 \times 3 \times 6}$ is used as input for Eq.~(\ref{eq:transf_atributo}) - (\ref{eq:ranking}), to obtain $\hat{\textbf{f}}$.  

We compare the ranking of features using prediction data $\hat{\textbf{f}}$ with five other rankings obtained from different approaches: \textbf{(1)} \textit{A benchmark $\textbf{f}^*$} -- this ranking is obtained using actual data of the years $ t =2013, \ldots, 2018$ to structure the tensor  $\mathcal{P}^* \in \mathbb{R}^{5 \times 3 \times 6}$ and used as input for PROMETHEE II extension,  Eq.~(\ref{eq:transf_atributo}) - (\ref{eq:ranking}). Note that $\textbf{f}^*$ is the $\hat{\textbf{f}}$ 's benchmark since, if the RLS algorithm provides good predicted values (low prediction error), $\hat{\textbf{f}}$ should be at least similar to $\textbf{f}^*$; \textbf{(2)} \textit{The classical approach used in MCDA, herein denoted by the \textit{current ranking} ($\textbf{g}^c$)} -- this ranking is obtained using the matrix $\textbf{H}  \in \mathbb{R}^{5 \times 3}$, which is structured with the data in $T = 33$ (the data in the year 2012), and used as input for the classical PROMETHEE II; \textbf{(3)} and  \textbf{(4)} \textit{Ranking obtained by the proposal in~\cite{campello2020adaptive}, that will be called prediction ranking ($\hat{\textbf{g}}$), and its reference ranking ($\textbf{g}^*$)}  -- to obtain the ranking ($\hat{\textbf{g}}$) the data is represented as a tensor $ \mathcal{H}  \in \mathbb{R}^{5 \times 3 \times 33}$, which feeds the algorithm  proposed in~\cite{campello2020adaptive}. The output is the ranking $\hat{\textbf{g}}$  for each year $t = 2013\ldots, 2018$. To obtain the reference ranking $\hat{\textbf{g}}$, the matrix $\textbf{H}  \in \mathbb{R}^{5 \times 3}$ is structured with the target values (actual data) for each year, followed by the application of the classical PROMETHEE II method; \textbf{(5)} \textit{The ranking obtained using the approach in~\cite{campello2023exploiting}, represented here by \textbf{g}} -- this ranking is obtained using a window of past data, specifically, the last six years' values ($t = 2007,\ldots, 2012$). Therefore, the tensor with past data is structured as $ \mathcal{B}  \in \mathbb{R}^{5 \times 3 \times 6}$, for data in $t = 2007,\ldots, 2012$, which is the input for PROMETHEE II extension,  Eq.~(\ref{eq:transf_atributo}) -- (\ref{eq:ranking}), and the output is the ranking \textbf{g}. All these rankings are shown in Table~\ref{tab:ranks}.

As can be seen in the first two rows of Table~\ref{tab:ranks}, the rankings provided by our approach ($\hat{\textbf{f}}$) and its main benchmark ($\textbf{f}^*$)  are the same ones, which means the prediction error did not affect the ranking. Besides, from Table~\ref{tab:ranks}, we observe that $\hat{\textbf{f}}$ is very different from $\textbf{g}^c$ since all alternatives are in a different position. Such differences can be explained with the aid of Table~\ref{tab:valores_feature_predicao_e_current}, which shows the tensor with features of the predicted values $\mathcal{S}  \in \mathbb{R}^{5 \times 3 \times 3}$ (the features of the predict signals), as well as the matrix of current values, $\textbf{H} \in \mathbb{R}^{5 \times 3}$. Let us focus on the values of the Alternatives 1 and 3 (which are ranked differently in $\hat{\textbf{f}}$ and $\textbf{g}^c$). From matrix \textbf{H}, one can note that $a_1$ outperforms $ a_3 $ for all criteria. In contrast, by analyzing the features obtained from the predicted signals, $a_3$ presents lower coefficient of variation for all criteria. Also, in the feature of $\mathcal{S}$ related to the average, $a_3$ is equal to $a_1$ in $c_2$. Thus, in the classical approach in MCDA, many essential decision elements are disregarded, even if they brings valuable information.

Moreover, it is interesting to note from Table~\ref{tab:ranks} that even when $\hat{\textbf{g}}$ and $\hat{\textbf{f}}$ are obtained using predicted values, the introduced approach brings additional information to the problem, which, in turn, may lead to different rankings, which may be more suitable in a given situation. For instance, $a_1$ ranks fourth in $\hat{\textbf{f}}$; however, it always ranks third in $\hat{\textbf{g}}$ and $\textbf{g}^*$. These results show in practice the discussion presented by Figure~\ref{fig:grafico}.

Continuing the analysis of Table~\ref{tab:ranks}, it is noteworthy that $\textbf{g} \neq \hat{\textbf{f}}$, as evidenced by the change in positions of $a_3$ and $a_1$. Table~\ref{tab:valores_feature_predicao_e_current} presents the tensor $\mathcal{T}$ containing the features calculated from the past samples; this tensor is obtained according to Eq. (2), but using  $ \mathcal{B}$ (tensor with past data) as input. From this table, we observe why $a_3$ and $a_1$ are in different positions in \textbf{g} and $\hat{\textbf{f}}$. From $\mathcal{S}$, we see that, for the feature given by the average, the value of $a_1$ is equal to the value of $a_3$ for $c_2$. In contrast, in $\mathcal{T}$, the value of  $a_3$ in $c_2$ is worse than the value of $a_1$ for $c_2$. Also, in $\mathcal{S}$, $a_3$ presents a lower coefficient of variation for all criteria than $a_1$. Instead, in $c_3$ from $a_3$ (in $\mathcal{T}$), the coefficient of variation is worse than in $a_1$. Therefore, in a six-year window, the signals show a change in their behavior. Such a behavior points out that the proposed approach was able to provide a meaningful even when there are abrupt changes in the average, tendency or variance.

\begin{table}
	\caption{Comparison of rankings obtained using NLMS and RLS as prediction methods, and the TOPSIS extension proposed by~\cite{campello2023exploiting} and PROMETHEE II extension as MCDA methods.  }\label{tab:ranks2}
	\begin{center}
		\scalebox{0.65}{%
			\begin{tabular}{lclccccc}
				\hline
				MCDA&Prediction&Ranking&1st &2nd &3rd &4th &5th \\\hline\hline
				\multirow{3}{*}{PROMETHEE II}& &Benchmark $\textbf{f}^*$		&$ a_5$& $ a_4 $&$ a_3 $&$ a_1 $&$ a_2$\\
				&RLS&Prediction $\hat{\textbf{f}}$		&$ a_5$& $ a_4 $&$ a_3 $&$ a_1 $&$ a_2$\\
				&LMS&Prediction $\hat{\textbf{f}}_{NLMS}$		&$ a_1$& $ a_2 $&$ a_3 $&$ a_4 $&$ a_5$\\\hline
				\hline
				\multirow{3}{*}{TOPSIS}& &Benchmark $\textbf{f}^{T*}$		&$ a_2$& $ a_5 $&$ a_3 $&$ a_1 $&$ a_4$\\
				&RLS&Prediction $\hat{\textbf{f}}^T$		&$ a_2$& $ a_5 $&$ a_3 $&$ a_1 $&$ a_4$\\
				&LMS&Prediction $\hat{\textbf{f}}_{NLMS}^T$		&$ a_3$& $ a_1 $&$ a_5 $&$ a_2 $&$ a_4$\\\hline
				
		\end{tabular}}
	\end{center}
\end{table}

We conducted further analysis of our proposal by elaborating Table~\ref{tab:ranks2}, which includes different methods of adaptive prediction and MCDA tensor aggregation. Rankings $\textbf{f}^*$ and  $\hat{\textbf{f}}$ were obtained as previously explained. To obtain ranking  $\hat{\textbf{f}}_{NLMS}$ we applied NLMS, instead of RLS, in the prediction step. Rankings $\textbf{f}^{T*}$, $\hat{\textbf{f}}^T$, and $\hat{\textbf{f}}_{NLMS}^T$ were generated using the TOPSIS extension proposed in~\cite{campello2023exploiting} instead of the PROMETHEE II extension. The input data used were the actual data, prediction data obtained from RLS, and prediction data obtained from the NLMS algorithm, respectively.  It's important to note that the ranking $\textbf{f}^{T*}$ is the benchmark for rankings  $\hat{\textbf{f}}^T$, and $\hat{\textbf{f}}_{NLMS}^T$. 

The first remark of Table~\ref{tab:ranks2} regards the prediction rankings generated using NLMS and RLS with both MCDA methods. The NLMS performs worse than RLS, as evidenced by the discrepancies $\hat{\textbf{f}}_{NLMS} \neq \textbf{f}^*$ and $\hat{\textbf{f}}_{NLMS}^T \neq \textbf{f}^{T*}$. The second observation is that both MCDA methods performed well in achieving the same ranking as their respective benchmarks, $\hat{\textbf{f}} = \textbf{f}^*$ and $\hat{\textbf{f}}^T = \textbf{f}^{T*}$, when RLS was used for the prediction data. Moreover, the obtained rankings using PROMETHEE II and TOPSIS differ,  $\textbf{f}^* \neq \textbf{f}^{T*}$, highlighting the importance of developing multiple methods for aggregating tensors, as previously discussed in the Introduction of this work.

\section{Conclusion}
\label{sec:conclusion}

This study presents an approach to support multiple criteria decisions using tensorial structures, an adaptive prediction algorithm, and a variant of the PROMETHEE II method to obtain a ranking of alternatives. The new approach was tested in a numerical experiment using real-life time series. We compare the ranking obtained with our proposal with the rankings obtained from other strategies. Such an analysis suggested that our proposal was able to capture temporal information of the prediction signals that is meaningful in many decisions processes. The proposed approach provided a proper ranking even in a nonstationary scenario in which the criteria present significant variations over a short time interval.

Moreover, we used different prediction methods for comparison, and RLS performed better than NLMS. Additionally, we compared the ranking obtained from the PROMETHEE II extension with the ranking obtained using a TOPSIS extension proposed in a previous study, and both rankings achieved their benchmark. However, they differed from each other due to their distinct characteristics.

\bibliography{bibliografia_doutorado.bib}

\begin{thebibliography}{10}
\expandafter\ifx\csname url\endcsname\relax
  \def\url#1{\texttt{#1}}\fi
\expandafter\ifx\csname urlprefix\endcsname\relax\def\urlprefix{URL }\fi
\expandafter\ifx\csname href\endcsname\relax
  \def\href#1#2{#2} \def\path#1{#1}\fi

\bibitem{marano2022decision}
S.~Marano, A.~H. Sayed, Decision-making algorithms for learning and adaptation
  with application to covid-19 data, Signal Processing 194 (2022) 108426.

\bibitem{hwang1981methods}
C.-L. Hwang, K.~Yoon, Methods for multiple attribute decision making, in:
  Multiple attribute decision making, Springer, 1981, pp. 58--191.

\bibitem{martins2021multidimensional}
M.~A. Martins, T.~V. Garcez, A multidimensional and multi-period analysis of
  safety on roads, Accident Analysis \& Prevention 162 (2021) 106401.

\bibitem{campello2023exploiting}
B.~S.~C. Campello, L.~T. Duarte, J.~M.~T. Romano, Exploiting temporal features
  in multicriteria decision analysis by means of a tensorial formulation of the
  topsis method, Computers \& Industrial Engineering 175 (2023) 108915.

\bibitem{zhang2017signal}
X.~P.~S. Zhang, F.~Wang, Signal processing for finance, economics, and
  marketing: Concepts, framework, and big data applications, IEEE Signal
  Processing Magazine 34~(3) (2017) 14--35.

\bibitem{agrahari2021prognosticating}
A.~Agrahari, P.~Singh, A.~Veer, A.~Singh, A.~Vidyarthi, B.~Khan,
  Prognosticating the effect on unemployment rate in the post-pandemic india
  via time-series forecasting and least squares approximation, Pattern
  Recognition Letters 152 (2021) 172--179.

\bibitem{dash2021intelligent}
S.~Dash, C.~Chakraborty, S.~K. Giri, S.~K. Pani, Intelligent computing on
  time-series data analysis and prediction of covid-19 pandemics, Pattern
  Recognition Letters 151 (2021) 69--75.

\bibitem{kandakoglu2019multicriteria}
A.~Kandakoglu, A.~Frini, S.~B. Amor, Multicriteria decision making for
  sustainable development: A systematic review, Journal of Multi-Criteria
  Decision Analysis 26~(5-6) (2019) 202--251.

\bibitem{frini2019mupom}
A.~Frini, S.~BenAmor, Mupom: A multi-criteria multi-period outranking method
  for decision-making in sustainable development context, Environmental Impact
  Assessment Review 76 (2019) 10--25.

\bibitem{mouhib2021tsmaa}
Y.~Mouhib, A.~Frini, {TSMAA-TRI}: A temporal multi-criteria sorting approach
  under uncertainty, Journal of Multi-Criteria Decision Analysis 28~(3-4)
  (2021) 185--199.

\bibitem{witt2021multi}
T.~Witt, M.~Klumpp, Multi-period multi-criteria decision making under
  uncertainty: A renewable energy transition case from germany, Sustainability
  13~(11) (2021) 6300.

\bibitem{wkatrobski2017temporal}
J.~Watrobski, W.~Salabun, G.~Ladorucki, The temporal supplier evaluation model
  based on multicriteria decision analysis methods, in: Asian Conference on
  Intelligent Information and Database Systems, Springer, 2017, pp. 432--442.

\bibitem{banamar2019interpolation}
I.~Banamar, An interpolation-based method for the time weighed vector
  elicitation in temporal promethee ii applications, International Journal of
  Multicriteria Decision Making 8~(1) (2019) 84--103.

\bibitem{campello2020adaptive}
B.~S. Campello, L.~T. Duarte, J.~M. Romano, Adaptive prediction of financial
  time-series for decision-making using a tensorial aggregation approach, in:
  IEEE International Conference on Acoustics, Speech and Signal Processing
  (ICASSP), 2020, pp. 5435--5439.

\bibitem{campello2022dealing}
B.~S.~C. Campello, L.~T. Duarte, J.~M.~T. Romano, Dealing with multi-criteria
  decision analysis in time-evolving approach using a probabilistic prediction
  method, Engineering Applications of Artificial Intelligence 116 (2022)
  105462.

\bibitem{kilmer2011factorization}
M.~E. Kilmer, C.~D. Martin, Factorization strategies for third-order tensors,
  Linear Algebra and its Applications 435~(3) (2011) 641--658.

\bibitem{chen2018tensor}
Y.~Chen, S.~Wang, Y.~Zhou, Tensor nuclear norm-based low-rank approximation
  with total variation regularization, IEEE Journal of Selected Topics in
  Signal Processing 12~(6) (2018) 1364--1377.

\bibitem{chen2020multi}
Y.~Chen, X.~Xiao, Y.~Zhou, Multi-view subspace clustering via simultaneously
  learning the representation tensor and affinity matrix, Pattern Recognition
  106 (2020) 107441.

\bibitem{brans2005promethee}
J.~P. Brans, B.~Mareschal, {PROMETHEE} methods, in: Multiple criteria decision
  analysis: state of the art surveys, Springer, 2005, pp. 163--186.

\bibitem{fu2023re}
G.~Fu, B.~Li, Y.~Yang, C.~Li, Re-ranking and {TOPSIS}-based ensemble feature
  selection with multi-stage aggregation for text categorization, Pattern
  Recognition Letters.

\bibitem{guitouni1998tentative}
A.~Guitouni, J.-M. Martel, Tentative guidelines to help choosing an appropriate
  {MCDA} method, European Journal of Operational Research 109~(2) (1998)
  501--521.

\bibitem{wkatrobski2016guideline}
J.~Watr{\'o}bski, J.~Jankowski, Guideline for mcda method selection in
  production management area, New Frontiers in Information and Production
  Systems Modelling and Analysis: Incentive Mechanisms, Competence Management,
  Knowledge-based Production (2016) 119--138.

\bibitem{zopounidis2002multicriteria}
C.~Zopounidis, M.~Doumpos, Multicriteria classification and sorting methods: a
  literature review, European Journal of Operational Research 138~(2) (2002)
  229--246.

\bibitem{romano2016unsupervised}
J.~M.~T. Romano, R.~Attux, C.~C. Cavalcante, R.~Suyama, Unsupervised signal
  processing: channel equalization and source separation, CRC Press, 2016.

\bibitem{feng2020enhancing}
F.~Feng, Z.~Xu, H.~Fujita, M.~Liang, Enhancing promethee method with
  intuitionistic fuzzy soft sets, International Journal of Intelligent Systems
  35~(7) (2020) 1071--1104.

\bibitem{behzadian2010promethee}
M.~Behzadian, R.~B. Kazemzadeh, A.~Albadvi, M.~Aghdasi, {PROMETHEE}: A
  comprehensive literature review on methodologies and applications, European
  journal of Operational research 200~(1) (2010) 198--215.

\end{thebibliography}

\end{document}